\documentclass[arxiv]{melba}

\usepackage{mwe} 
\usepackage{graphicx}

\usepackage[utf8]{inputenc}
\usepackage[T1]{fontenc}
\usepackage{graphicx}
\usepackage{subcaption} 
\usepackage{diagbox}
\usepackage{bm}
\usepackage{amsmath}
\usepackage{amssymb}
\usepackage{multirow}
\usepackage{enumitem}
\usepackage{amsmath,amsfonts}
\newcolumntype{L}[1]{>{\raggedright\arraybackslash}p{#1}}
\newcolumntype{C}[1]{>{\centering\arraybackslash}p{#1}}
\newcolumntype{R}[1]{>{\raggedleft\arraybackslash}p{#1}}

\melbaid{YYYY:NNN}  
\doi{10.59275/j.melba.2024-AAAA}
\melbaauthors{Name1 and Name2}  
\email{siyuan.mei@fau.de}
\volume{2}
\firstpageno{1337}  
\melbayear{YYYY}  
\datesubmitted{yyyy-m1-d1}  
\datepublished{yyyy-m2-d2}  

\melbaspecialissue{Medical Imaging with Deep Learning (MIDL) 2020}
\melbaspecialissueeditors{Marleen de Bruijne, Tal Arbel, Ismail Ben Ayed, Hervé Lombaert}

\ShortHeadings{BigReg: An Efficient Registration Pipeline for High-Resolution X-Ray and Light-Sheet Fluorescence Microscopy}{Mei, Fan, Thies, Gu, Wagner, Aust, Erceg, Mirzaei, Neag, Sun, Huang, Maier}

\title{BigReg: An Efficient Registration Pipeline for High-Resolution X-Ray and Light-Sheet Fluorescence Microscopy}

\author{
	\firstname Siyuan \surname Mei\aff{1},
	\name Fuxin Fan\aff{1},
        \name Mareike Thies\aff{1},
         \name Mingxuan Gu\aff{1},
        \name Fabian Wagner\aff{1},
        \name Oliver Aust\aff{2},
        \name Ina Erceg\aff{3},
        \name Zeynab Mirzaei\aff{4},
        \name Georgiana Neag\aff{2},
        \name Yipeng Sun\aff{1},
        \name Yixing Huang\aff{5},
        \name Andreas Maier\aff{1},
}

\affiliations{
	\num 1 \addr Pattern Recognition Lab, Friedrich-Alexander University Erlangen-Nuremberg, Erlangen, Germany \\
	\num 2 \addr Department of Internal Medicine 3 - Rheumatology and Immunology, University Hospital Erlangen, Erlangen, Germany \\
	\num 3 \addr Fraunhofer Institute for Ceramic Technologies and Systems - IKTS, Forcheim, Germany \\
    \num 4 \addr Institute for Nanotechnology and Correlative Microscopy - INAM GmbH, Forcheim, Germany \\
    \num 5 \addr Institute of Medical Technology, Peking University, Beijing, China \\
}

\abstract{
	Recently, X-ray microscopy (XRM) and light-sheet fluorescence microscopy (LSFM) have emerged as pivotal tools in preclinical research, particularly for studying bone remodeling diseases such as osteoporosis. These modalities offer micrometer-level resolution, and their integration allows for a complementary examination of bone microstructures which is essential for analyzing functional changes. However, registering high-resolution volumes from these independently scanned modalities poses substantial challenges, especially in real-world and reference-free scenarios. This paper presents BigReg, a fast, two-stage pipeline designed for large-volume registration of XRM and LSFM data. The first stage involves extracting surface features and applying two successive point cloud-based methods for coarse alignment. The subsequent stage refines this alignment using a modified cross-correlation technique, achieving precise volumetric registration. Evaluations using expert-annotated landmarks and augmented test data demonstrate that BigReg approaches the accuracy of landmark-based registration with a landmark distance (LMD) of 8.36\,\textmu m\,$\pm$\,0.12\,\textmu m and a landmark fitness (LM fitness) of 85.71\%\,$\pm$\,1.02\%. Moreover, BigReg can provide an optimal initialization for mutual information-based methods which otherwise fail independently, further reducing LMD to 7.24\,\textmu m\,$\pm$\,0.11\,\textmu m and increasing LM fitness to 93.90\%\,$\pm$\,0.77\%. Ultimately, key microstructures, notably lacunae in XRM and bone cells in LSFM, are accurately aligned, enabling unprecedented insights into the pathology of osteoporosis.}


\begin{document}

\twocolumn[\maketitle]

\section{Introduction}

\begin{figure*}[tb]
    \centering
    \begin{subfigure}[b]{0.2\textwidth}
        \centering
        \includegraphics[width=0.5\textwidth]{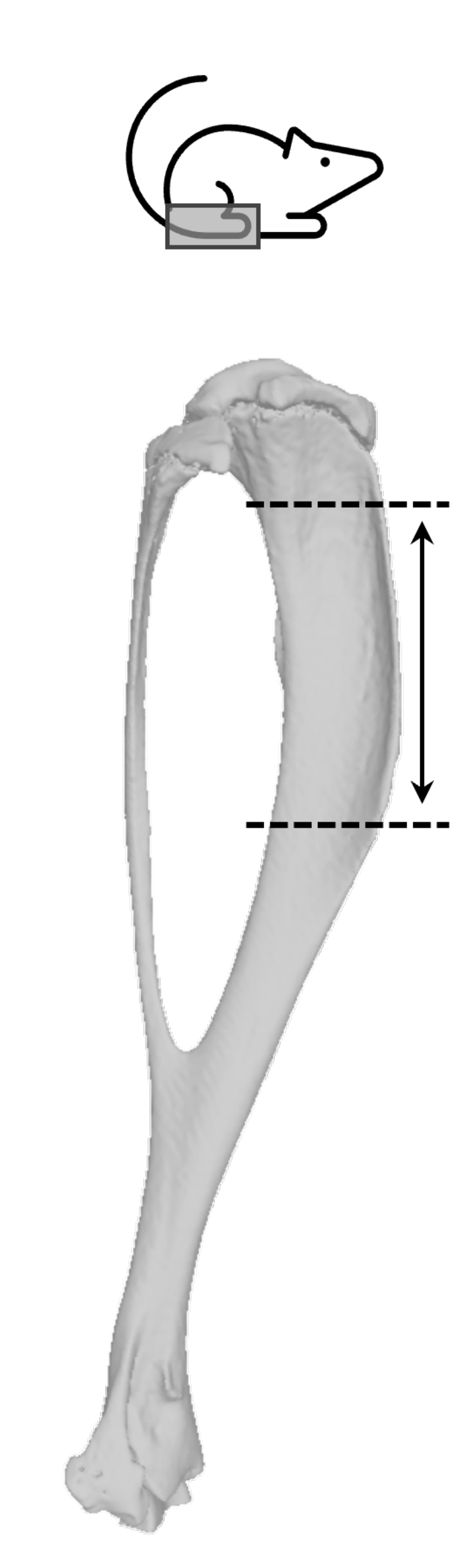}
         \caption*{(a) Mouse tibia sample} 
    \end{subfigure}
    \hspace{0.02\textwidth}
    \begin{subfigure}[b]{0.32\textwidth}
        \centering
        \includegraphics[width=\textwidth]{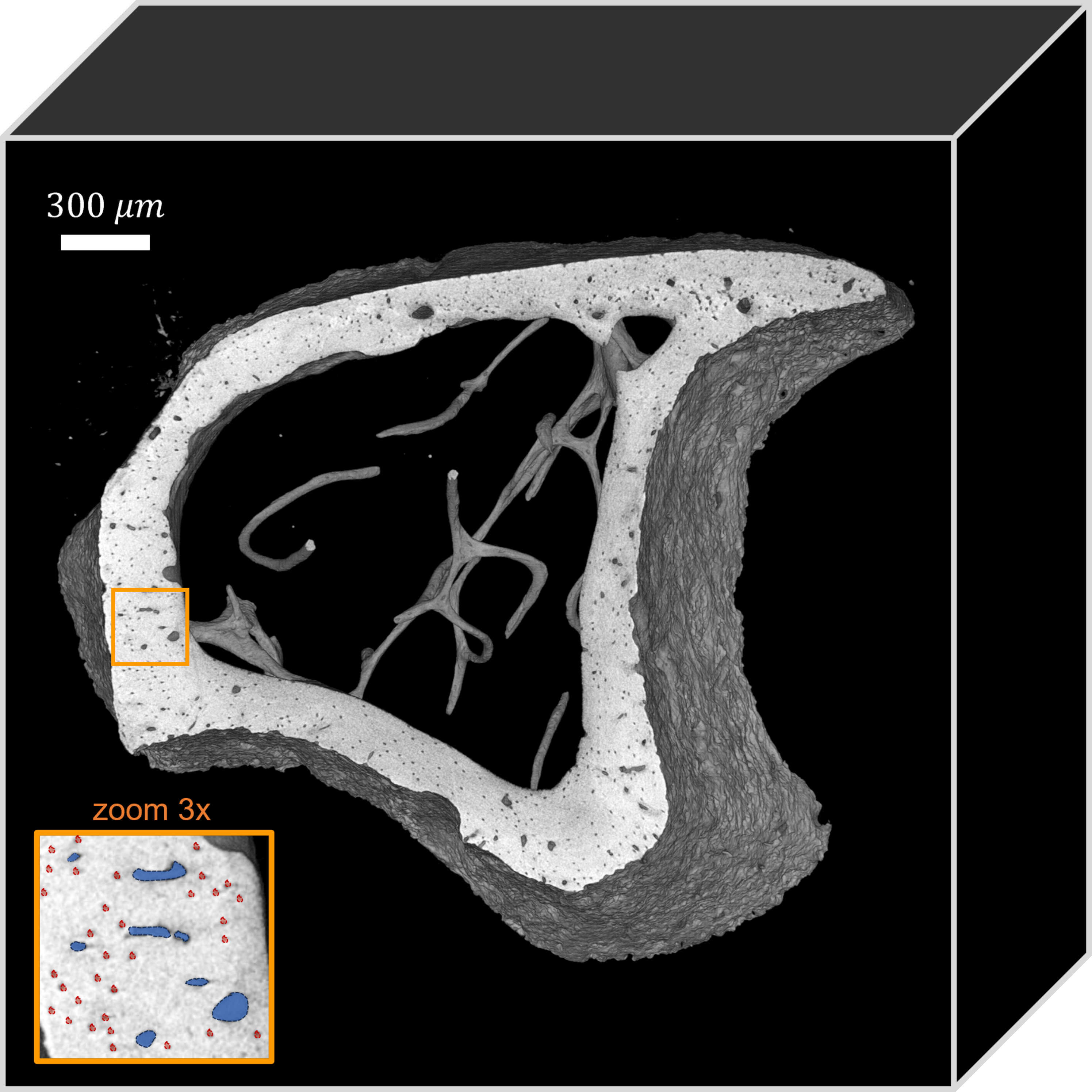}
        \caption*{(b) XRM volume} %
        \label{xrm}
    \end{subfigure}
    \hspace{0.02\textwidth} %
    \begin{subfigure}[b]{0.32\textwidth}
        \centering
        \includegraphics[width=\textwidth]{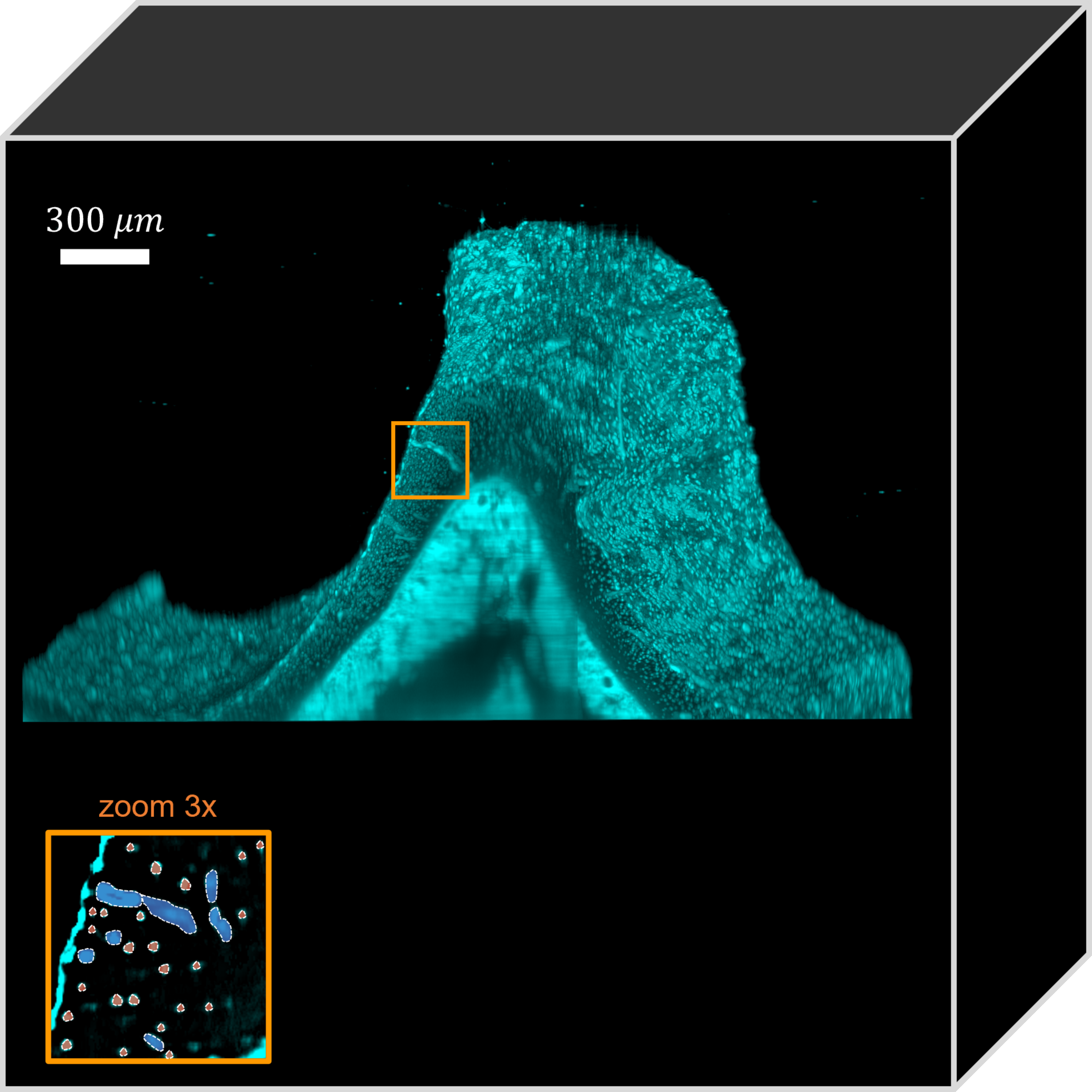}
         \caption*{(c) LSFM volume} 
         \label{lsfm}
    \end{subfigure}

    \caption{Visualization of (a) a mouse tibia sample used as the scanning subject and 3D rendered volume for (b) XRM data and (c) LSFM data. Both volumes are grayscale but shown in different colors, and their voxel sizes and shapes are unified to 1.42\,$\times$\,1.42\,$\times$\,1.42\,\textmu m$^3$ and 2048\,$\times$\,2048\,$\times$\,800, respectively. While the XRM scan reconstructs the complete bone structure, the LSFM scan omits the lower part of the bone cross-section due to increased scattering along the detection depth. In the highlighted subregions within the orange boxes, which are magnified threefold, the XRM volume features vessel canals (blue labels) and lacunae (red labels), whereas the LSFM volume shows vessels (blue labels) and cell nuclei (red labels).}
    \label{fig:rendered volume}
\end{figure*}


\enluminure{O}{steoporosis} (OP) is the predominant cause of fragility fractures among the elderly, characterized by age-related bone loss and deterioration of the bone microarchitecture~\citep{clynes2020epidemiology}. The latest systematic review reported a global osteoporosis prevalence of 19.7\%, with postmenopausal women exhibiting the highest rate at 27.4\%~\citep{xiao2022global}. Pathologically, this disease arises from imbalanced bone remodeling governed by various bone cells, such as osteocytes~\citep{raggatt2010cellular}. These osteocytes reside within microscopic cavities called lacunae, which are a few micrometers in size and interconnected through tiny canals~\citep{buenzli2015quantifying, peyrin2014micro}. In preclinical studies, ovariectomized rodent (OVX) models, established by removing the ovaries from female mice or rats, are commonly used to mimic the estrogen deficiency-induced bone loss seen in postmenopausal women~\citep{yousefzadeh2020ovariectomized}. Conventional biological investigations monitor trabecular bone density in OVX models using micro-computed tomography (CT)~\citep{xiong2021evaluation, akhter2021high}. However, two significant limitations of micro-CT prevent deep exploration into the mechanisms behind micro-architectural deterioration at the cellular scale: the resolution around 20-50 \textmu m is insufficient to resolve the lacunae~\citep{thies2022calibration, longo2017comparison}, and its monomodality cannot capture the interaction between the lacunae and the osteocytes they host. To address these shortcomings, recent research has utilized X-ray microscopy (XRM)~\citep{langer20163d} and light-sheet fluorescence microscopy (LSFM)~\citep{thai2024using, xiao2024reduction} to generate volumetric data from mouse bones at single-micrometer precision. These advanced modalities provide complementary 3D views, with XRM revealing detailed bone morphology and LSFM highlighting cellular and vascular structures within the bones (see details in Fig.~\ref{fig:rendered volume}). Early efforts combining these modalities supported by expert observation have unveiled the significance of small transcortical vessels~\citep{gruneboom2019network}. Nonetheless, this manual approach is not only laborious but also falls short for finer structures which are essential for innovative osteoporosis metrics, such as the ratio of empty lacunae to those containing bone cells. To advance this research and fully exploit the potential of these scans, the development of an automated algorithm for precise registration of XRM and LSFM volumes has become a priority. 

In this paper, we propose BigReg, a highly efficient two-stage registration pipeline for high-resolution XRM and LSFM volumes, as depicted in Fig.~\ref{fig:pipline}. Given a volume pair as input, the first stage extracts surface points and then performs a global-to-local point cloud registration, providing a coarse alignment. Crucially, this initial alignment is refined in the second stage through an efficient search for the remaining transformation in the Fourier domain based on a modified cross-correlation method. Our results suggest that BigReg can match the performance of manual landmark-based methods and is robust against significant misregistrations. The main contributions are summarized as follows:
\begin{itemize}
\item[$\bullet$] This is the first work that successfully performs automatic registration between high-resolution XRM and LSFM volumes while attaining micrometer-scale accuracy.
\item[$\bullet$] The combination of point cloud-based and cross-correlation methods considers both surface features and full image contents, achieving coarse to fine registration within two stages.
\item[$\bullet$] The entire pipeline is both memory- and computation-efficient, achieved by minimizing the data amount through surface feature extraction in the first stage and reducing computational load via Fourier transformation in the second stage.
\item[$\bullet$] Last but not least, we elucidate the critical role of the initial transformation in MI-based methods and demonstrate how BigReg can serve as an optimal initializer, thus enhancing the quest for ultimate registration precision
\end{itemize}

\section{Related Works}

\begin{figure*}[tb]
\centering\includegraphics[width=\textwidth]{pipeline.pdf}
\caption{Illustration of BigReg pipeline. From left to right, the moving XRM volume needs to be rigidly transformed to align with the fixed LSFM volume. Stage 1 extracts surface features and performs two successive point cloud-based registration procedures to offer a coarse alignment $\textbf{T}_1$. Stage 2 refines this alignment using masked normalized cross-correlation (MNCC) in the Fourier domain to achieve finer volumetric registration $\textbf{T}_2$. Finally, the registration result is automatically obtained by combining these two transformations.} \label{fig:pipline}
\end{figure*}

Image registration refers to aligning one image (the moving image) with another image (the fixed image) by identifying the appropriate spatial transform. This field has undergone significant exploration over the past four decades, particularly in different medical imaging modalities~\citep{saiti2020application, viergever2016survey, bharati2022deep, sengupta2022survey}. Progress has been notable in modalities such as CT to magnetic resonance imaging (MRI) or within multi-channel MRIs. Focusing on 3D rigid transformation (3D rotation and translation), we categorize existing methods into three groups and discuss their challenges in managing the complex modalities addressed in this study.

\subsection{Feature-based methods} Feature-based methods rely on extracting salient features common to different images, such as surface contours~\citep{song2017registration, sinko20183d, harada2008optimal} and anatomical landmarks~\citep{strasters1997anatomical, pennec2000landmark}. As an easily available feature without the need for fiducial markers, surface-based approaches perform reliably by finding point-to-point correspondences using matching techniques like iterative closest points (ICP)~\citep{icp, yang2015go}. Registration using these geometrical features benefits from efficient processing time due to the reduced data amount. However, they restrict themselves to a subset of the available volumetric information which does not meet the fine-grained registration accuracy required by microstructure analysis~\citep{savva2016geometry}. 

\subsection{Intensity-based methods} In contrast, intensity-based methods take full image contents into account and iteratively optimize transformation parameters to reach a maximum similarity measure between two images~\citep{rahunathan2005image, avants2009advanced, modat2014global, klein2009elastix}. The choice of metrics depends on the modality differences, with mutual information (MI)~\citep{rahunathan2005image, sengupta2022survey} commonly used across different modalities and cross-correlation (CC)~\citep{avants2008symmetric} within the same modalities. Despite their popularity, these methods are infeasible for large-sized volumes with billions of voxels, as each iteration step executes an intermediate transformation interpolating the entire volume which is both computationally intensive and time-consuming. Moreover, issues like partial overlap and extreme misregistrations can lead to higher similarity scores for incorrect alignments, causing convergence failures of the non-convex function~\citep{sengupta2022survey}. 

\subsection{Learning-based methods}

\begin{figure*}[htb]
    \centering
    \hspace{0.3cm}
    \begin{subfigure}[b]{0.35\textwidth}
        \centering
        \includegraphics[width=\textwidth]{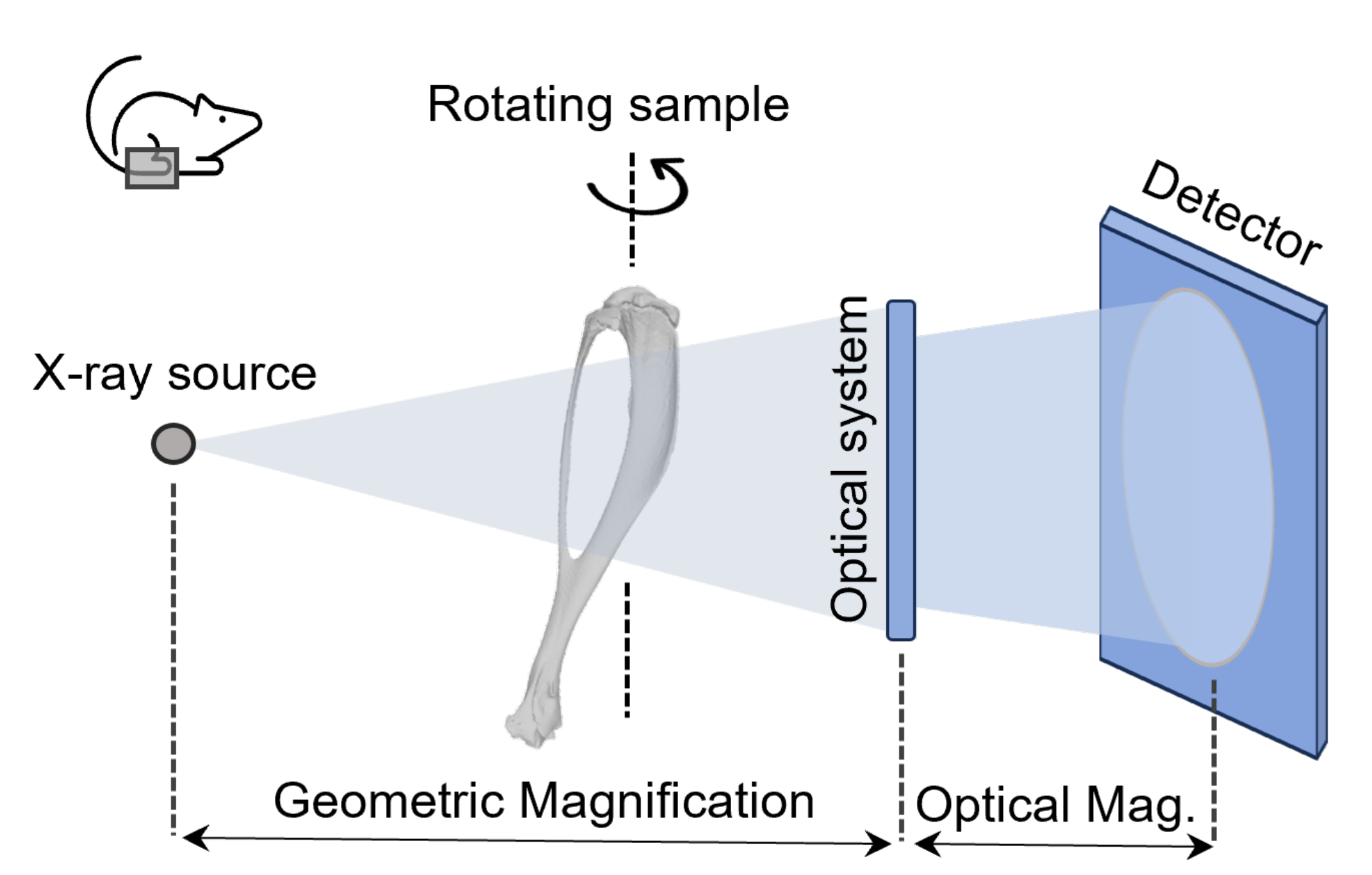}
        \caption*{(a) XRM scanning} %
    \end{subfigure}
    \hspace{1.cm} %
    \begin{subfigure}[b]{0.4\textwidth}
        \centering
        \includegraphics[width=\textwidth]{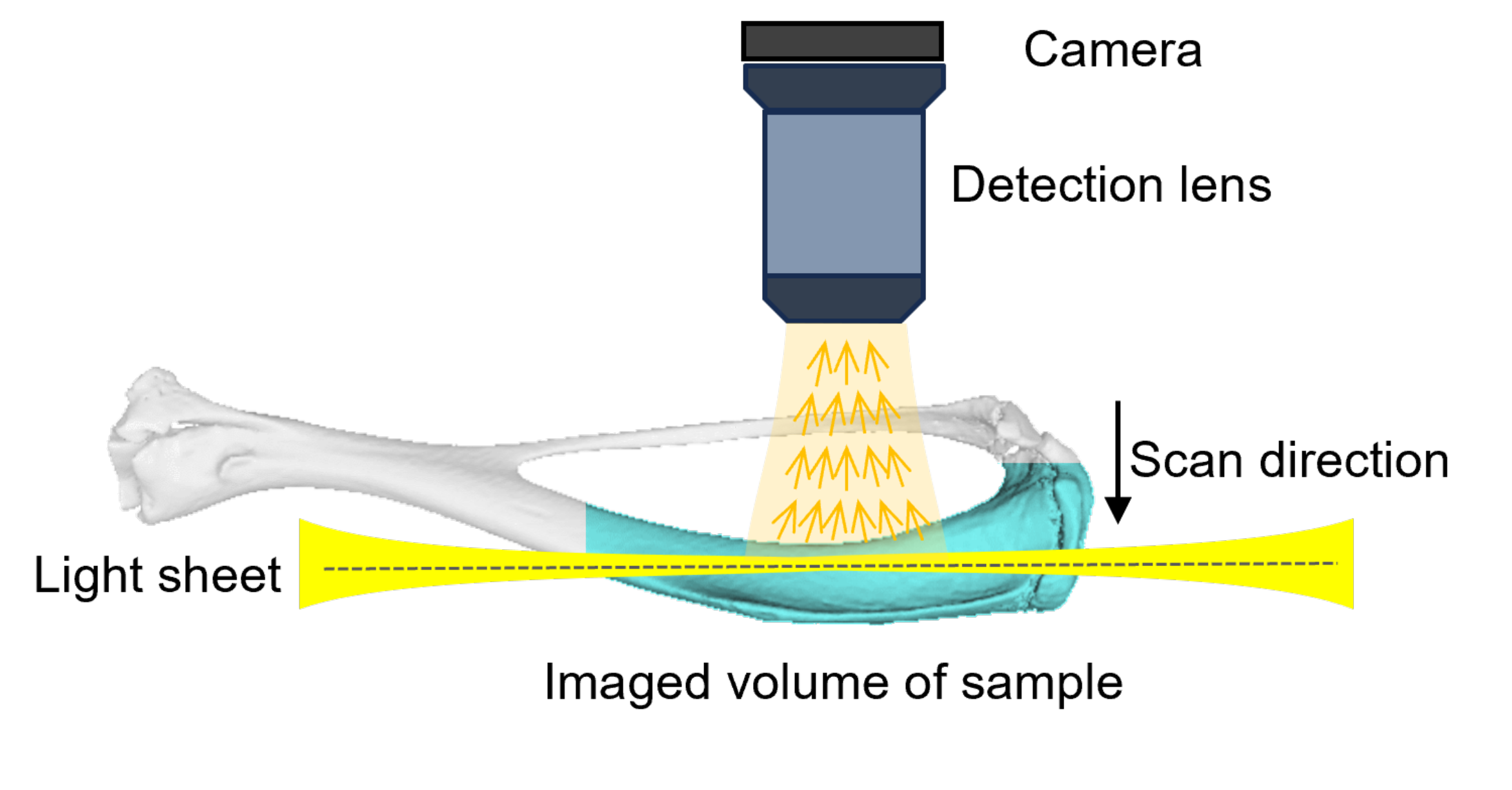}
         \caption*{(b) LSFM scanning} 
    \end{subfigure}

    \caption{Illustration of scanning procedure of (a) XRM and (b) LSFM. Specifically, XRM captures high-resolution projection images through the two-stage magnification architecture; LSFM records horizontal slices illuminated by the corresponding thin light sheet.}
    \label{fig:imaging}
\end{figure*}

Recent deep learning-based (DL) methods~\citep{bharati2022deep, wang2023robust, de2019deep, lee2019image} unsupervisedly register images using deep convolutional layers to extract 3D features, subsequently regressing transformation parameters through linear layers or a center-of-mass layer~\citep{ma2020volumetric}. These methods utilize loss functions similar to those in classical approaches~\citep{wang2023robust}. Unfortunately, for clinical applications involving unique biological samples, the available data are often too sparse to support data-driven learning. Meanwhile, maintaining full resolution in 3D data is crucial as downscaling would dissolve critical tiny structures. Yet, the generic GPU memory is inadequate to handle such large datasets, further impeding the training of deep neural networks. Therefore, DL methods are not considered in this paper.

\section{Methods}
\label{sec:2}

\subsection{Data description and preprocessing} \label{sec:2.1} 

XRM and LSFM utilize distinct imaging protocols, as illustrated in Fig.~\ref{fig:imaging}. In XRM scanning, X-ray beams pass through a rotating sample, which are subsequently converted into visible light via an optical system. A detector captures the attenuated photons to create 2D projections which are then used for 3D reconstruction. With different X-ray absorption rates, bone structures are imaged at higher intensity against soft tissues' markedly lower intensity, revealing blood vessel canals and lacunae. Conversely, LSFM is an optical imaging method relying on fluorescent markers. To highlight functional soft tissues within the bones, vessels, bone cells, and bone marrow are labeled with specific fluorescent stains, whereas bones are rendered transparent~\citep{santi2011light}. During scanning, a light sheet illuminates a fluorescence plane, capturing one slice of the sample area with a camera. The scanner collects a series of slices by moving the light sheet along the scan direction, forming a 3D volume. Therefore, LSFM images capture fluorescing soft tissues with high intensity, offering a complementary view to those obtained by XRM.

Table~\ref{xrm scan parameter} and Table~\ref{lsfm scan parameter} respectively summarize the key parameters of XRM and LSFM scan settings used in our experiments, with the registration-related image attributes highlighted in bold. To standardize the data for registration, we undertake several preprocessing steps for each volume pair: Firstly, the LSFM volume is resliced and rescaled to match the XRM scan's coordinate system and resolution of 1.42 \textmu m in all dimensions; Secondly, the distal bone segments comprising 800 slices are selected to focus on medically most relevant regions; Finally, both volumes are zero-padded to achieve uniform dimensions of 2048\,$\times$\,2048\,$\times$\,800 and normalized to [0, 255]. Fig.~\ref{fig:rendered volume} shows an example of the processed volume pair, with target microstructures labeled in the magnified subregions. Compared to the entire bone structure in the XRM reconstructed volume, the lower portion of the bone cross-section in the LSFM volume has not been acquired due to significant scattering artifacts. In the following registration process, we define XRM as the moving volume ($\bm{V}_{M}$) and LSFM as the fixed volume ($\bm{V}_{F}$).

\begin{table*}[tb]
    \centering
    \begin{minipage}{.5\linewidth}
        \centering
        \begin{tabular}{L{4.5cm}R{3.5cm}}
        \toprule
        \multicolumn{2}{c}{Device: Xradia 620 Versa microscope (Carl Zeiss)}                    \\ 
        \midrule
        X-ray source voltage & 60 kVp \\
         X-ray source current & 108 \textmu A \\
         Number of projection images & 1401 \\
         Angular range & $202^\circ$ (short scan) \\
         Detector shape & 2038\,$\times$\,2038\\
         Detector pixel size & 11.0 \textmu m\\
         Source–isocentre distance & 16.5 mm\\
         Isocentre–detector distance & 16.0 mm\\
         Magnification (geometrical/optical/total) & 1.97/4/7.88 \\
        \textbf{Reconstructed volume shape}     &  1997\,$\times$\,2038\,$\times$\,2014 \\
        \textbf{(height\,$\times$\,width\,$\times$\,slices)} & \\
        \textbf{Voxel size}         & 1.42\,$\times$\,1.42\,$\times$\,1.42 \textmu $\text{m}^3$  \\
        \bottomrule
        \end{tabular}
        \caption{XRM scan settings}\label{xrm scan parameter}
    \end{minipage}%
    \hspace{.02\linewidth}
    \begin{minipage}{.45\linewidth}
        \centering

        \begin{tabular}{L{3.5cm}R{3.5cm}}
        \toprule
        \multicolumn{2}{c}{Device: Ultramicroscope II (Miltenyi Biotec)}  \\ \midrule
         Illumination & Bidirectional \\
         Number of light sheets & 1 \\
         Light sheet width & 20 mm \\
         Zoom & Mono zoom \\
         Zoom ratio & 2.0\\
         Objective lenses  & 2 \\
         Total magnification & 4.0 \\
         Numerical aperture (NA) & 0.5 \\
         NA for the sheet thickness & 0.16 \\
        \textbf{Volume shape}               &  2048\,$\times$\,2048\,$\times$\,504 \\
         \textbf{(height\,$\times$\,width\,$\times$\,slices)} & \\
        \textbf{Voxel size}                       &  1.51\,$\times$\, 1.51\,$\times$\, 4.00\,\textmu $\text{m}^3$ \\
        \bottomrule
        \end{tabular}
        \caption{LSFM scan settings}\label{lsfm scan parameter}
    \end{minipage}
\end{table*}


\subsection{Stage 1: surface point cloud registration}
\label{sec: 2.2}
The first stage of the registration pipeline starts with extracting surface point clouds from two volumes. We begin with segmenting the bone shape at a low threshold ($\bm{V}_{M}>5$ and $\bm{V}_{F}>0$). A series of binary morphological operations—comprising binary closing, hole filling, and outlining—is then performed on each slice to delineate the surface contours. The coordinates of all surface points are extracted to form point clouds. As illustrated in the upper middle part of Fig.~\ref{fig:pipline}, the moving point cloud $\bm{P}_{M}\in\mathbb{Z}^{3\times N_M}$ is rendered in yellow, whereas the fixed point cloud $\bm{P}_{F}\in\mathbb{Z}^{3\times N_F}$ appears in cyan, where $N_M$ and $N_F$ represent the number of points in $\bm{P}_{M}$ and $\bm{P}_{F}$, respectively. After generating these two surface point clouds, we initially align their centers to minimize the initial displacement and prepare for the subsequent two substages of registration.


\subsubsection{Stage 1.1: FPFH RANSAC}

In substage 1.1, we employ the fast point feature histogram (FPFH) descriptor~\citep{rusu2009fast} to map each point of $\bm{P}_{M}$ and $\bm{P}_{F}$ into a 33-dimensional feature vector, resulting in $\bm{P}_{M}^{FPFH}\in\mathbb{R}^{33\times N_M}$ and $\bm{P}_{F}^{FPFH}\in\mathbb{R}^{33\times N_F}$. The FPFH feature captures the geometric relationship histogram between a source point and its specified number of neighbors within a defined spherical range~\citep{rusu2009fast}. Two hyperparameters are used to determine the search space: $n_{FPFH}$, the maximum number of neighbors, and $r_{FPFH}$, the radius of the sphere. Therefore, the corresponding point pairs can be found by querying the nearest neighbors in the global FPFH feature space. We apply the random sampling consensus (RANSAC) algorithm~\citep{fischler1981random} to iteratively fit a rigid transformation model between the two point clouds, following these steps: 
\begin{enumerate}[label=(\roman*)]
    \item Randomly pick three points from $\bm{P}_{M}^{FPFH}$ and detect their corresponding points from $\bm{P}_{F}^{FPFH}$ using nearest neighbor search in the FPFH feature space, facilitated by a K-D Tree~\citep{kdtree}.
    \item For these matched point pairs, compute a hypothesized transformation matrix $\bm{T}\in\mathbb{R}^{4\times4}$ based on Umeyama least-squares estimation~\citep{umeyama1991least}.
    \item Apply $\bm{T}$ to $\bm{P}_{M}$ and determine the number of inliers, where inliers are considered as those neighboring point pairs from two point clouds with distances below a predefined threshold of $d_{FPFH}$.
    \item Repeat the above steps until the maximum number of iterations, $iter_{FPFH}$, is reached, and then return the transformation matrix with the highest number of inliers.
\end{enumerate}

Since the correspondence detection is executed for all points, this substage is able to provide a global surface registration without prior alignment, yielding an initial transformation matrix $\bm{T}_{1.1}$.

\subsubsection{Stage 1.2: point-to-plane ICP} 

The FPFH RANSAC algorithm provides a suboptimal solution for registration due to its reliance on random sampling. Following a widely recognized workflow in point cloud registration~\citep{open3d, pcl}, we integrate the point-to-plane ICP~\citep{p2licp} to achieve local refinement. Given a pre-alignment $\bm{T}_{1.1}$, this method presumes the closest points as correspondences and iteratively refines the registration to achieve the tightest fit, as detailed in the following steps:
\begin{enumerate}[label=(\roman*)]
    \item Apply the current transformation $\bm{T}$ to $\bm{P}_{M}$, resulting in the transformed point cloud $\bm{P}^\prime_{M}$. 
    \item Identify the set of closest point pairs $\mathcal{K}\{(\bm{p}_M^{\prime(i)}, \bm{p}_F^{(i)})\},\,i=0,...,N_F-1$ from $\bm{P}^\prime_{M}$ and $\bm{P}_{F}$, considering a maximum correspondence distance $d_{ICP}$.
    \item Update $\bm{T}$ by minimizing the following objective function with the Gauss-Newton optimizer
    \begin{equation}
        \mathcal{L}(\bm{T})=\sum_{(\bm{p}_M^{\prime(i)}, \bm{p}_F^{(i)})\in\mathcal{K}}{[\bm{n}^{(i)}\cdot(\bm{p}_F^{(i)}-\bm{T}\bm{p}_M^{\prime(i)})]^2},
    \end{equation}
    where $\bm{n}^{(i)}$ is the surface normal vector at point $\bm{p}_F^{(i)}$.
    \item Repeat the steps until the designated maximum number of iterations ($iter_{ICP}$) is reached and determine the final transformation.
\end{enumerate}

At the end of this process, the refined transformation $\bm{T}_{1.2}$ that accurately aligns the two point clouds is derived. The output of the first stage is updated as $\bm{T}_1=\bm{T}_{1.2}$ that transforms the moving volume to $\bm{V}_{M}^{1}=a(\bm{V}_{M},\bm{T}_1)$, where $a(\cdot)$ represents the rigid transformation with bicubic interpolation.

\subsection{Stage 2: volumetric registration}
\label{sec: 2.3}
Following the surface registration, stage two aims to correct the residual misalignments of microstructures within the bone. Re-observing the image contents in Fig.~\ref{fig:rendered volume}, vessel canals and lacunae which manifest as tiny holes with nearly zero intensity in XRM volumes, contrast sharply with the bright, high-intensity representations of stained vessels and cells in LSFM volumes. These opposite intensity features can be matched by inverting the voxel intensity of XRM volume, yielding $\overline{\bm{V}_{M}^{1}}$. To eliminate the influence of the background outside the bone, binary bone masks $\bm{h}_F$ and $\bm{h}_M$ are generated for $\bm{V}_{F}$ and $\overline{\bm{V}_{M}^{1}}$ respectively using thresholding at zero followed by hole-filling operations. In addition, we apply the unsharp mask algorithm~\citep{morishita1988unsharp} with a Gaussian kernel of 5 and weight of 0.8 for LSFM to reduce low-frequency scattering and enhance feature contrast, producing $S(\bm{V}_{F})$. Four cropped slices from $S(\bm{V}_{F})$ with mask $\bm{h}_F$ and $\overline{\bm{V}_{M}^{1}}$ with mask $\bm{h}_M$ are shown in the lower middle part of Fig.~\ref{fig:pipline}. For brevity, we redefine $S(\bm{V}_{F})\odot\bm{h}_F$ as $\bm{V}_1$ and $\overline{\bm{V}_{M}^{1}}\odot\bm{h}_M$ as $\bm{V}_2$, where $\odot$ is element-wise multiplication.

Now, the high-intensity voxels in both $\bm{V}_1$ and $\bm{V}_2$ present the same semantic in bones. An optimal registration is achieved by finding a transformation $\bm{T}_2$ that maximizes the sum of the product of $\bm{V}_1$ and $a(\bm{V}_2, \bm{T}_2)$. Given that large displacement and rotation have been compensated during the surface registration phase, the remaining transformation can be approximated as a tight 3D translation, where $\bm{T}_2\approx[\Delta x, \Delta y, \Delta z]$. We conduct an exhaustive search for $\bm{T}_2$ using masked normalized cross-correlation (MNCC), formulated as
\begin{equation}
    MNCC(\bm{T}_2)= \frac{\sum{[(\bm{V}_1-\mu_1)(a(\bm{V}_2, \bm{T}_2)-\mu_2)]}}{\sqrt{\sum{(\bm{V}_1-\mu_1)}^2}\sqrt{\sum{(a(\bm{V}_2, \bm{T}_2)-\mu_2)}^2}},
    \label{eq: mncc}
\end{equation}
where $\mu_1$ and $\mu_2$ represent the mean value of $\bm{V}_1$ and $a(\bm{V}_2, \bm{T}_2)$ within the overlapping regions of $\bm{h}_F$ and $a(\bm{h}_M, \bm{T}_2)$, respectively. 

Like the cross-correlation in the spatial domain has its dual in the Fourier domain~\citep{bracewell1966fourier}, MNCC also benefits from a Fourier transform variant~\citep{padfield2011masked}, enabling efficient computation across all potential transformations using the fast Fourier transform (FFT). This approach is vital for large volumes as it bypasses the need for direct transformations in the spatial domain, significantly accelerating the registration process. The outcome of the FFT MNCC operation generates a comprehensive search space of the same size as the original volume, as illustrated in the lower middle section of Fig.~\ref{fig:pipline}. Each coordinate represents a relative translation ($[0,0,0]$ at the center), and the intensity corresponds to the MNCC score within the range of $[-1, 1]$. The optimal transformation is identified as the offset from the center to the location with the peak correlation score. Consequently, the overall transformation, combining both stages, is given as $\bm{T}_{overall}=\bm{T}_2\bm{T}_1$.


\section{Experimental setup}

\subsection{Data} 

We used a total of seven ex-vivo mouse tibial samples, each independently scanned by XRM and LSFM to create corresponding volume pairs. For the experiments, we augmented the dataset by applying a random pre-transformation to each registration run with rotations ranging from [-180$^\circ$, 180$^\circ$] and translations from [0, 100] on the xy-plane. To expedite processing, all intermediate transformations utilized linear interpolation. All experiments were conducted on identical hardware configurations, specifically an Intel i5-13500 CPU with 64 GB RAM.

\subsection{Hyperparameters} 

The first stage of our pipeline depends on hyperparameters affecting the surface point cloud. We determined the most robust set of hyperparameters through extensive grid searching. For stage 1.1, the optimal parameters identified were: $iter_{FPFH} = 2.5 \times 10^6$, $r_{FPFH} = 200$, $n_{FPFH} = 400$, and $d_{FPFH} = 30$. For stage 1.2, the optimal hyperparameters were $d_{ICP} = 16$ and $iter_{ICP} = 2000$.

\subsection{Evaluation} 

To evaluate our method, we manually annotated 50 pairs of landmarks for each volume pair, focusing on corresponding anatomical microstructures under the supervision of biological experts. All landmark coordinates were distributed widely in 3D space, each at a distance of more than 100 \textmu m from its nearest neighbors. We employed RANSAC with $10^5$ iterations and three correspondences to derive a transformation matrix for these defined landmark pairs. This landmark-based registration (LMR) serves as our ground truth (GT). In addition, four metrics are used for quantitative evaluation: 

\begin{figure*}[tb]
    \centering\includegraphics[width=0.9\textwidth]{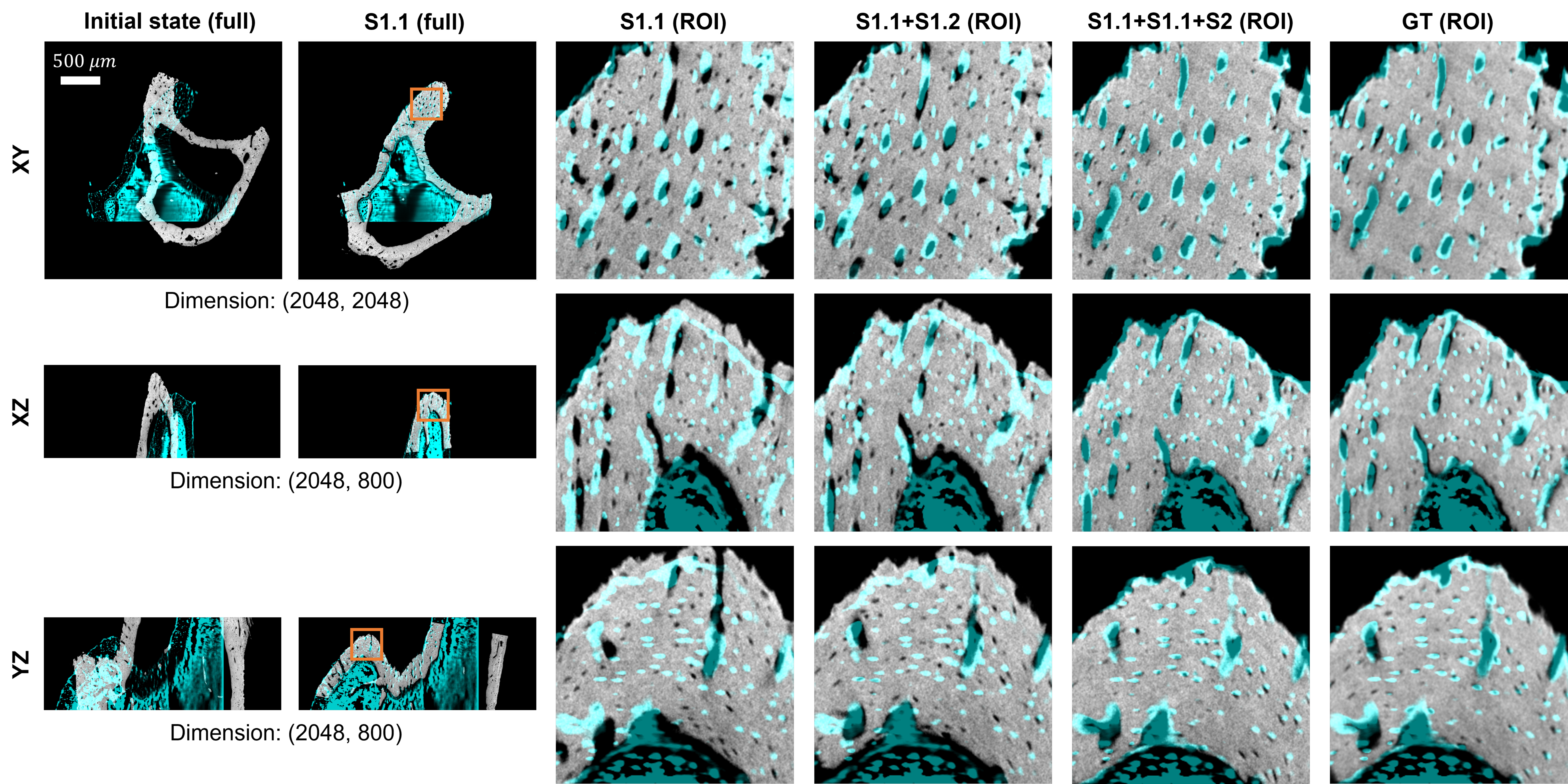}
    \caption{Visualization of the registration process in BigReg. Rows display central slices from different planes of the same sample. Columns one and two show the entire views at the initial state and S1.1, respectively, with specified dimensions. The subsequent three columns highlight eightfold magnified regions of interest (ROIs) from the $256\times256$ orange boxes through successive registration stages. The last column displays the GT result.} \label{qualitative_bigreg}
\end{figure*}

\begin{table*}[tb]

    \centering
    \begin{tabular}{c|c|c|c|c|c}
        \hline
         BigReg Component &  Only S1.1 &  Only S1.2 & S1.1+S1.2 & S1.1+S2 & S1.1+S1.2+S2 \\
         \hline
         LMD [\textmu m] ($\downarrow$) & 43.27$\pm$16.09 & 894.84$\pm$468.08 & 25.91$\pm$0.52 & 16.08$\pm$5.38 & 8.36$\pm$0.12\\
         LM fitness [$\%$] ($\uparrow$)& 7.23$\pm$0.09 & 0.03$\pm$0.00 & 27.65$\pm$0.01 & 47.67$\pm$17.58 & 85.71$\pm$1.02\\
         Rotation error [$^\circ$] ($\downarrow$) & 2.52$\pm$0.64 & 88.65$\pm$55.16 & 0.71$\pm$0.01 & 2.25$\pm$0.86 & 0.71$\pm$0.01\\
         Translation error [\textmu m] ($\downarrow$) & 73.06$\pm$20.88  & 2758.81$\pm$1535.46  &  40.11$\pm$0.76 & 65.12$\pm$31.26 & 13.90$\pm$0.35 \\
         \hline
    \end{tabular}
    \caption{Ablation study results for different pipeline configurations, reflecting mean and standard deviation across all samples when tested 100 times with random initial transformations.}\label{table1}
\end{table*}

\begin{itemize}[label={--},leftmargin=*]
     \item \textbf{Landmark distance (LMD)} measures the average distance between moved XRM landmarks and fixed LSFM landmarks. 
     \item \textbf{Landmark fitness (LM fitness)} quantifies the inlier ratio of the landmarks, where the inlier threshold is set to 12 \textmu m.
     \item \textbf{Rotation error} refers to the minimum angular difference needed to align the estimated and GT rotation in space.
     \item \textbf{Translation error} is the Euclidean distance between the estimated and GT translation vectors.
\end{itemize}




\section{Results}

\subsection{Ablation study}

To assess the contribution of each component within the BigReg pipeline, we conducted an ablation study under various configurations: ``only stage 1.1 (S1.1)'', ``only stage 1.2 (S1.2)'', ``S1.1+S1.2'', ``S1.1+S2'', and the complete pipeline ``S1.1+S1.2+S2''. Table~\ref{table1} presents averaged results across all samples, each tested 100 times with random initial transformations to gauge robustness. As evidenced in the second and third columns of the table, results from ``only S1.1'' are absolutely better than those from ``only S1.2''. Adhering to a global-to-local registration strategy, S1.2 notably enhances the performance initiated by S1.1. From the fourth to the last column, we can observe that the full BigReg pipeline yields the highest performance across all metrics and demonstrates substantial robustness with the lowest standard deviation, whereas removing either S1.2 or S2 will degrade the performance. These findings are further visualized in Fig.~\ref{qualitative_bigreg}, which illustrates the progressive alignment of target microstructures throughout successive stages of the BigReg process, culminating in precise alignment after the full pipeline execution.


\begin{figure*}[tb]
    \centering\includegraphics[width=\textwidth]{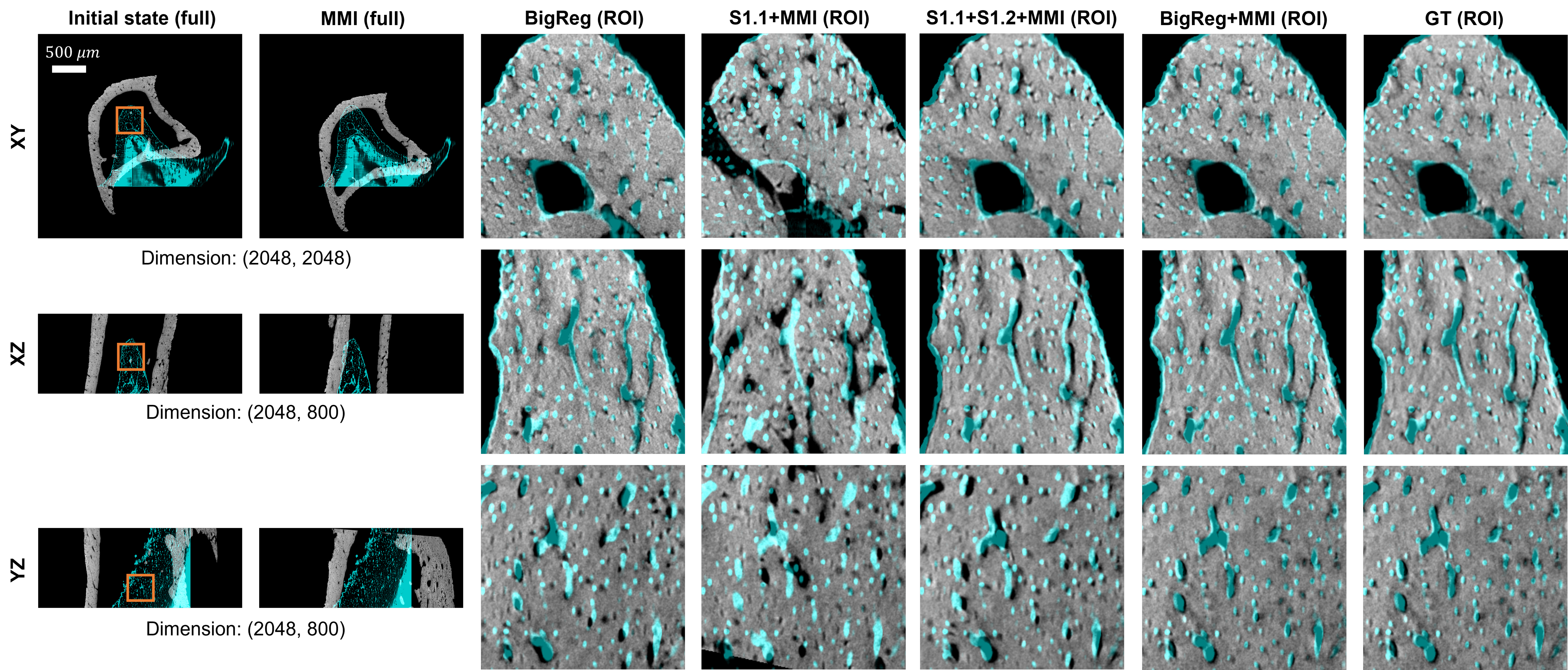}
    \caption{Qualitative results on an extremely misregistered example. Rows display central slices from different planes of the same sample. The first column shows the initial state in full view. Except for the second column which presents MMI's failed result, subsequent columns highlight eightfold magnified ROI from the $256\times256$ orange boxes, corresponding to the applied methods.} \label{Qualitative results}
\end{figure*}

\begin{table*}[tb]

    \centering
    \begin{tabular}{c|c|c|c|c|c}
        \hline
         Methods &  MMI & BigReg & S1.1+MMI & S1.1+S1.2+MMI  &  BigReg+MMI \\  
         \hline
         LMD [\textmu m] ($\downarrow$) &  >1000 &8.36$\pm$0.12 & 19.12$\pm$20.54 & 8.50$\pm$5.46 & 7.24$\pm$0.11  \\    
         LM fitness [$\%$] ($\uparrow$)& 0.00 & 85.71$\pm$1.02 & 83.14$\pm$20.78 & 93.78$\pm$4.11 & 93.90$\pm$0.77 \\   
         Rotation error [$^\circ$] ($\downarrow$) & >120 & 0.71$\pm$0.01 & 0.63$\pm$0.60&0.32$\pm$0.10 & 0.34$\pm$0.04 \\  
         Translation error [\textmu m] ($\downarrow$) & >2000  &  13.90$\pm$0.35 & 29.26$\pm$34.32 & 10.63$\pm$5.02 &  11.29$\pm$1.44 \\  
         \hline
    \end{tabular}
    \caption{Quantitative results compared with different baseline methods. The statistical mean and standard deviation were calculated by randomly transforming each sample 100 times. Note that an intrinsic LMD of 6.10\,\textmu m exists in the GT result.}\label{tab: quanti}
\end{table*}

\subsection{Comparison against baselines}

\subsubsection{Baselines}

We compare BigReg against several baseline methods:

\begin{itemize}[label={--},leftmargin=*]
     \item \textbf{LMR}: As described earlier, LMR using 50 manually annotated landmarks is treated as the ground truth (GT), although it has an average LMD of 6.10 µm due to minor annotation and measurement errors.
     \item \textbf{Mattes mutual information (MMI)}: MMI~\citep{mattes2001nonrigid} is a classical but most universal baseline. We utilize the implementation in ANTs, a state-of-the-art software framework for medical image registration~\citep{avants2009advanced}. Specifically, we provide the bone masks to normalize the partial overlap and employ a multi-resolution pyramid strategy, registering volumes progressively at three different resolutions: 0.25x, 0.5x, and finally at full resolution.
     \item \textbf{Initializers+MMI}: We observe that MMI tends to be significantly affected by large initial misregistration. To investigate the effectiveness of MMI, we provide different-level initializations from BigReg to drive it. Here, we separately allocated ``S1.1'', ``S1.1+S1.2'', and the full BigReg as initializers to MMI. To the best of our knowledge, this idea has never been mentioned by other existing approaches. Therefore, it's also regarded as an extension of our proposed method.
    %
\end{itemize}




\subsubsection{Qualitative and quantitative results}

Fig.~\ref{Qualitative results} displays results on a challenging misregistration case. From column two, it is evident that MMI alone fails to adequately register the volume. In contrast, BigReg alone aligns the microstructures accurately, except for minor differences in the YZ plane compared to GT (visual difference can be observed between columns three and seven of Fig.~\ref{Qualitative results}). Subsequent results demonstrate that MMI's performance is highly sensitive to initialization. While MMI with either ``S1.1'' or ``S1.1+S1.2'' performs worse than standalone BigReg (columns four and five), combining the full BigReg with MMI achieves the best alignment, virtually indistinguishable from GT (column six).

The same observation is reflected in the quantitative results in Table~\ref{tab: quanti}. MMI deviates significantly from GT across all metrics. In contrast, our proposed BigReg demonstrates robust results, approximating GT with an LMD of 8.36\,\textmu m\,$\pm$\,0.12\,\textmu m and an LM fitness of 85.71\%\,$\pm$\,1.02\%. In column five, it is noteworthy that ``S1.1+S1.2+MMI'' slightly outperforms standalone BigReg in the mean values except for LMD, but it exhibits lower robustness with consistently higher standard deviations. Finally, the extended BigReg with MMI continues to improve performance, achieving the lowest LMD of 7.24\,\textmu m\,$\pm$\,0.11\,\textmu m and the highest LM fitness of 93.90\%\,$\pm$\,0.77\%, along with minimal rotation and translation errors.

\section{Discussion}

The presented pipeline solves micrometer-level registration within two steps, with both conducted in an efficient way considering the enormous volume size. According to the ablative experiments, each stage plays an indispensable role within the pipeline. S1.1 lays the groundwork by providing a global transformation, while S1.2 and S2 further enhance performance and robustness. Specifically, omitting S2 leads to remaining translation errors (as shown in Table~\ref{table1}, column four), and the removal of S1.2 increases the instability of the method (Table~\ref{table1}, column five).

Furthermore, our observations reveal that the optimization process for MMI tends to terminate prematurely, often yielding inadequate transformation parameters. This issue is evident in the second column of Fig.~\ref{Qualitative results}, where the moving volume adjusts only minimally, primarily the distance between the centers of the two volumes. We attribute this to severe partial overlap and misalignment, which misleadingly elevate the MI value and cause the optimization gradients to incorrectly direct the registration process. However, with proper initialization, MMI regains its effectiveness. Thus, the integration of BigReg and MMI represents a synergistic approach, where BigReg provides optimal initialization and MMI addresses the rotation errors that BigReg’s second stage might overlook. 

In addition, it is meaningful to mention that BigReg avoids optimizing for the entire volume, which allows for completing the task within approximately ten minutes—two minutes for the registration algorithm and eight minutes for one interpolated transformation. Although MMI does not fully account for the registration time due to its early stopping, one can still imagine that the multiple intermediate transformations could take a significant amount of time, measured in hours.   

Looking ahead, we plan to explore the development of an auto-adaptive point cloud-based method to refine the parameter-dependent stage one, aiming to broaden the applicability of our approach. Although our current focus is on XRM and LSFM data, the fundamental principles of BigReg could potentially be generalized to other complex medical datasets.



\section{Conclusion}

We present an automatic two-stage registration pipeline, BigReg, tailored for high-resolution XRM and LSFM volumes. This innovative pipeline achieves performance comparable to manually annotated landmark-based registration methods. Through extensive experimentation, we demonstrated that while MMI-based methods typically struggle with such challenging data, they can be revitalized by integration with BigReg, thus forming an enhanced version of the pipeline. Most importantly, with the successful fusion of two image modalities, it is for the first time possible to conduct large-scale studies quantifying the proportion of lacunae that contain osteocytes as a potential new biomarker for bone remodeling diseases.

\acks{This work was supported by the European Research Council (ERC Grant No. 810316). Sample preparation was conducted at University Hospital Erlangen, and data collection at Fraunhofer Institute for Ceramic Technologies and Systems (IKTS) and Institute for Nanotechnology and Correlative Microscopy (INAM).}

%
\ethics{All animal experiments in this study were performed in accordance with German legal and regulatory guidelines and were approved by the local animal ethics committee of the Regierung von Mittelfranken (TS-12/2015). Moreover, the procedures conformed to the guidelines of the Federation of European Laboratory Animal Science Associations and the ARRIVE (Animal Research: Reporting of In Vivo Experiments) guidelines. Female C57BL/6JRj mice (14 weeks old) obtained from the University Hospital Erlangen were used in the research. All mice were anesthetized and subjected to bilateral ovariectomy or a sham operation. For tissue sample preparation, the mice were euthanized by CO2. Tibiae were relieved from muscle tissue and post-fixed in 4\% PFA/PBS (pH 7.4) for 4 hours at 4–8$^\circ$C with gentle shaking. Tissue fixation was followed by 100\% ethanol dehydration. }

\coi{We declare we don't have conflicts of interest.}

\data{The dataset is available from the corresponding author upon reasonable request.}

\bibliography{sample}

\begin{thebibliography}{47}
\providecommand{\natexlab}[1]{#1}
\providecommand{\url}[1]{\texttt{#1}}
\expandafter\ifx\csname urlstyle\endcsname\relax
  \providecommand{\doi}[1]{doi: #1}\else
  \providecommand{\doi}{doi: \begingroup \urlstyle{rm}\Url}\fi

\bibitem[Akhter and Recker(2021)]{akhter2021high}
MP~Akhter and RR~Recker.
\newblock High resolution imaging in bone tissue research-review.
\newblock \emph{Bone}, 143:\penalty0 115620, 2021.

\bibitem[Avants et~al.(2008)Avants, Epstein, Grossman, and Gee]{avants2008symmetric}
Brian~B Avants, Charles~L Epstein, Murray Grossman, and James~C Gee.
\newblock Symmetric diffeomorphic image registration with cross-correlation: evaluating automated labeling of elderly and neurodegenerative brain.
\newblock \emph{Medical image analysis}, 12\penalty0 (1):\penalty0 26--41, 2008.

\bibitem[Avants et~al.(2009)Avants, Tustison, Song, et~al.]{avants2009advanced}
Brian~B Avants, Nick Tustison, Gang Song, et~al.
\newblock Advanced normalization tools (ants).
\newblock \emph{Insight j}, 2\penalty0 (365):\penalty0 1--35, 2009.

\bibitem[Bentley(1975)]{kdtree}
Jon~Louis Bentley.
\newblock Multidimensional binary search trees used for associative searching.
\newblock \emph{Communications of the ACM}, 18\penalty0 (9):\penalty0 509--517, 1975.

\bibitem[Besl and McKay(1992)]{icp}
Paul~J Besl and Neil~D McKay.
\newblock Method for registration of 3-d shapes.
\newblock In \emph{Sensor fusion IV: control paradigms and data structures}, volume 1611, pages 586--606. Spie, 1992.

\bibitem[Bharati et~al.(2022)Bharati, Mondal, Podder, and Prasath]{bharati2022deep}
Subrato Bharati, M~Mondal, Prajoy Podder, and VB~Prasath.
\newblock Deep learning for medical image registration: A comprehensive review.
\newblock \emph{arXiv preprint arXiv:2204.11341}, 2022.

\bibitem[Bracewell and Kahn(1966)]{bracewell1966fourier}
Ron Bracewell and Peter~B Kahn.
\newblock The fourier transform and its applications.
\newblock \emph{American Journal of Physics}, 34\penalty0 (8):\penalty0 712--712, 1966.

\bibitem[Buenzli and Sims(2015)]{buenzli2015quantifying}
Pascal~R Buenzli and Natalie~A Sims.
\newblock Quantifying the osteocyte network in the human skeleton.
\newblock \emph{Bone}, 75:\penalty0 144--150, 2015.

\bibitem[Chen and Medioni(1992)]{p2licp}
Yang Chen and G{\'e}rard Medioni.
\newblock Object modelling by registration of multiple range images.
\newblock \emph{Image and vision computing}, 10\penalty0 (3):\penalty0 145--155, 1992.

\bibitem[Clynes et~al.(2020)Clynes, Harvey, Curtis, Fuggle, Dennison, and Cooper]{clynes2020epidemiology}
Michael~A Clynes, Nicholas~C Harvey, Elizabeth~M Curtis, Nicholas~R Fuggle, Elaine~M Dennison, and Cyrus Cooper.
\newblock The epidemiology of osteoporosis.
\newblock \emph{British medical bulletin}, 133\penalty0 (1):\penalty0 105--117, 2020.

\bibitem[De~Vos et~al.(2019)De~Vos, Berendsen, Viergever, Sokooti, Staring, and I{\v{s}}gum]{de2019deep}
Bob~D De~Vos, Floris~F Berendsen, Max~A Viergever, Hessam Sokooti, Marius Staring, and Ivana I{\v{s}}gum.
\newblock A deep learning framework for unsupervised affine and deformable image registration.
\newblock \emph{Medical image analysis}, 52:\penalty0 128--143, 2019.

\bibitem[Fischler and Bolles(1981)]{fischler1981random}
Martin~A Fischler and Robert~C Bolles.
\newblock Random sample consensus: a paradigm for model fitting with applications to image analysis and automated cartography.
\newblock \emph{Communications of the ACM}, 24\penalty0 (6):\penalty0 381--395, 1981.

\bibitem[Gr{\"u}neboom et~al.(2019)Gr{\"u}neboom, Hawwari, Weidner, Culemann, M{\"u}ller, Henneberg, Brenzel, Merz, Bornemann, Zec, et~al.]{gruneboom2019network}
Anika Gr{\"u}neboom, Ibrahim Hawwari, Daniela Weidner, Stephan Culemann, Sylvia M{\"u}ller, Sophie Henneberg, Alexandra Brenzel, Simon Merz, Lea Bornemann, Kristina Zec, et~al.
\newblock A network of trans-cortical capillaries as mainstay for blood circulation in long bones.
\newblock \emph{Nature metabolism}, 1\penalty0 (2):\penalty0 236--250, 2019.

\bibitem[Harada et~al.(2008)Harada, Kim, Tan, Ishikawa, and Yamamoto]{harada2008optimal}
Kouhei Harada, Hyoungseop Kim, Joo~Kooi Tan, Seiji Ishikawa, and Akiyoshi Yamamoto.
\newblock Optimal registration method based on icp algorithm from head ct and mr image sets.
\newblock In \emph{2008 International Conference on Control, Automation and Systems}, pages 1268--1271. IEEE, 2008.

\bibitem[Klein et~al.(2009)Klein, Staring, Murphy, Viergever, and Pluim]{klein2009elastix}
Stefan Klein, Marius Staring, Keelin Murphy, Max~A Viergever, and Josien~PW Pluim.
\newblock Elastix: a toolbox for intensity-based medical image registration.
\newblock \emph{IEEE transactions on medical imaging}, 29\penalty0 (1):\penalty0 196--205, 2009.

\bibitem[Langer and Peyrin(2016)]{langer20163d}
Max Langer and Francoise Peyrin.
\newblock 3d x-ray ultra-microscopy of bone tissue.
\newblock \emph{Osteoporosis International}, 27:\penalty0 441--455, 2016.

\bibitem[Lee et~al.(2019)Lee, Oktay, Schuh, Schaap, and Glocker]{lee2019image}
Matthew~CH Lee, Ozan Oktay, Andreas Schuh, Michiel Schaap, and Ben Glocker.
\newblock Image-and-spatial transformer networks for structure-guided image registration.
\newblock In \emph{Medical Image Computing and Computer Assisted Intervention--MICCAI 2019: 22nd International Conference, Shenzhen, China, October 13--17, 2019, Proceedings, Part II 22}, pages 337--345. Springer, 2019.

\bibitem[Longo et~al.(2017)Longo, Salmon, and Ward]{longo2017comparison}
Amanda~B Longo, Phil~L Salmon, and Wendy~E Ward.
\newblock Comparison of ex vivo and in vivo micro-computed tomography of rat tibia at different scanning settings.
\newblock \emph{Journal of Orthopaedic Research}, 35\penalty0 (8):\penalty0 1690--1698, 2017.

\bibitem[Ma et~al.(2020)Ma, Gupta, and Sabuncu]{ma2020volumetric}
Tianyu Ma, Ajay Gupta, and Mert~R Sabuncu.
\newblock Volumetric landmark detection with a multi-scale shift equivariant neural network.
\newblock In \emph{2020 IEEE 17th International Symposium on Biomedical Imaging (ISBI)}, pages 981--985. IEEE, 2020.

\bibitem[Mattes et~al.(2001)Mattes, Haynor, Vesselle, Lewellyn, and Eubank]{mattes2001nonrigid}
David Mattes, David~R Haynor, Hubert Vesselle, Thomas~K Lewellyn, and William Eubank.
\newblock Nonrigid multimodality image registration.
\newblock In \emph{Medical imaging 2001: image processing}, volume 4322, pages 1609--1620. Spie, 2001.

\bibitem[Modat et~al.(2014)Modat, Cash, Daga, Winston, Duncan, and Ourselin]{modat2014global}
Marc Modat, David~M Cash, Pankaj Daga, Gavin~P Winston, John~S Duncan, and S{\'e}bastien Ourselin.
\newblock Global image registration using a symmetric block-matching approach.
\newblock \emph{Journal of medical imaging}, 1\penalty0 (2):\penalty0 024003--024003, 2014.

\bibitem[Morishita et~al.(1988)Morishita, Yamagata, Okabe, Yokoyama, and Hamatani]{morishita1988unsharp}
Koichi Morishita, Shimbu Yamagata, Tetsuo Okabe, Tetsuo Yokoyama, and Kazuhiko Hamatani.
\newblock Unsharp masking for image enhancement, December~27 1988.
\newblock US Patent 4,794,531.

\bibitem[Padfield(2011)]{padfield2011masked}
Dirk Padfield.
\newblock Masked object registration in the fourier domain.
\newblock \emph{IEEE Transactions on image processing}, 21\penalty0 (5):\penalty0 2706--2718, 2011.

\bibitem[Pennec et~al.(2000)Pennec, Ayache, and Thirion]{pennec2000landmark}
Xavier Pennec, Nicholas Ayache, and Jean-Philippe Thirion.
\newblock Landmark-based registration using features identified through differential geometry, 2000.

\bibitem[Peyrin et~al.(2014)Peyrin, Dong, Pacureanu, and Langer]{peyrin2014micro}
Fran{\c{c}}oise Peyrin, Pei Dong, Alexandra Pacureanu, and Max Langer.
\newblock Micro-and nano-ct for the study of bone ultrastructure.
\newblock \emph{Current osteoporosis reports}, 12:\penalty0 465--474, 2014.

\bibitem[Raggatt and Partridge(2010)]{raggatt2010cellular}
Liza~J Raggatt and Nicola~C Partridge.
\newblock Cellular and molecular mechanisms of bone remodeling.
\newblock \emph{Journal of biological chemistry}, 285\penalty0 (33):\penalty0 25103--25108, 2010.

\bibitem[Rahunathan et~al.(2005)Rahunathan, Stredney, Schmalbrock, and Clymer]{rahunathan2005image}
Smriti Rahunathan, Don Stredney, P~Schmalbrock, and Bradley~D Clymer.
\newblock Image registration using rigid registration and maximization of mutual information.
\newblock In \emph{13th Annu. Med. Meets Virtual Reality Conf}, 2005.

\bibitem[Rusu and Cousins(2011)]{pcl}
Radu~Bogdan Rusu and Steve Cousins.
\newblock 3d is here: Point cloud library (pcl).
\newblock In \emph{2011 IEEE international conference on robotics and automation}, pages 1--4. IEEE, 2011.

\bibitem[Rusu et~al.(2009)Rusu, Blodow, and Beetz]{rusu2009fast}
Radu~Bogdan Rusu, Nico Blodow, and Michael Beetz.
\newblock Fast point feature histograms ({FPFH}) for {3D} registration.
\newblock In \emph{2009 IEEE international conference on robotics and automation}, pages 3212--3217. IEEE, 2009.

\bibitem[Saiti and Theoharis(2020)]{saiti2020application}
Evdokia Saiti and Theoharis Theoharis.
\newblock An application independent review of multimodal 3d registration methods.
\newblock \emph{Computers \& Graphics}, 91:\penalty0 153--178, 2020.

\bibitem[Santi(2011)]{santi2011light}
Peter~A Santi.
\newblock Light sheet fluorescence microscopy: a review.
\newblock \emph{Journal of Histochemistry \& Cytochemistry}, 59\penalty0 (2):\penalty0 129--138, 2011.

\bibitem[Savva et~al.(2016)Savva, Economopoulos, and Matsopoulos]{savva2016geometry}
Antonis~D Savva, Theodore~L Economopoulos, and George~K Matsopoulos.
\newblock Geometry-based vs. intensity-based medical image registration: A comparative study on 3d ct data.
\newblock \emph{Computers in biology and medicine}, 69:\penalty0 120--133, 2016.

\bibitem[Sengupta et~al.(2022)Sengupta, Gupta, and Biswas]{sengupta2022survey}
Debapriya Sengupta, Phalguni Gupta, and Arindam Biswas.
\newblock A survey on mutual information based medical image registration algorithms.
\newblock \emph{Neurocomputing}, 486:\penalty0 174--188, 2022.

\bibitem[Sinko et~al.(2018)Sinko, Kamencay, Hudec, and Benco]{sinko20183d}
Martin Sinko, Patrik Kamencay, Robert Hudec, and Miroslav Benco.
\newblock 3d registration of the point cloud data using icp algorithm in medical image analysis.
\newblock In \emph{2018 ELEKTRO}, pages 1--6. IEEE, 2018.

\bibitem[Song et~al.(2017)Song, Jiang, Yang, Wang, and Zhang]{song2017registration}
Zhiying Song, Huiyan Jiang, Qiyao Yang, Zhiguo Wang, and Guoxu Zhang.
\newblock A registration method based on contour point cloud for 3d whole-body pet and ct images.
\newblock \emph{BioMed Research International}, 2017\penalty0 (1):\penalty0 5380742, 2017.

\bibitem[Strasters et~al.(1997)Strasters, Little, Buurman, Hill, and Hawkes]{strasters1997anatomical}
Karel~C Strasters, John~A Little, Johannes Buurman, Derek~LG Hill, and David~J Hawkes.
\newblock Anatomical landmark image registration: validation and comparison.
\newblock In \emph{International Conference on Computer Vision, Virtual Reality, and Robotics in Medicine}, pages 161--170. Springer, 1997.

\bibitem[Thai et~al.(2024)Thai, Fuller-Jackson, and Ivanusic]{thai2024using}
Jenny Thai, John-Paul Fuller-Jackson, and Jason~J Ivanusic.
\newblock Using tissue clearing and light sheet fluorescence microscopy for the three-dimensional analysis of sensory and sympathetic nerve endings that innervate bone and dental tissue of mice.
\newblock \emph{Journal of Comparative Neurology}, 532\penalty0 (1):\penalty0 e25582, 2024.

\bibitem[Thies et~al.(2022)Thies, Wagner, Huang, Gu, Kling, Pechmann, Aust, Gr{\"u}neboom, Schett, Christiansen, et~al.]{thies2022calibration}
Mareike Thies, Fabian Wagner, Yixing Huang, Mingxuan Gu, Lasse Kling, Sabrina Pechmann, Oliver Aust, Anika Gr{\"u}neboom, Georg Schett, Silke Christiansen, et~al.
\newblock Calibration by differentiation--self-supervised calibration for x-ray microscopy using a differentiable cone-beam reconstruction operator.
\newblock \emph{Journal of Microscopy}, 287\penalty0 (2):\penalty0 81--92, 2022.

\bibitem[Umeyama(1991)]{umeyama1991least}
Shinji Umeyama.
\newblock Least-squares estimation of transformation parameters between two point patterns.
\newblock \emph{IEEE Transactions on Pattern Analysis \& Machine Intelligence}, 13\penalty0 (04):\penalty0 376--380, 1991.

\bibitem[Viergever et~al.(2016)Viergever, Maintz, Klein, Murphy, Staring, and Pluim]{viergever2016survey}
Max~A Viergever, JB~Antoine Maintz, Stefan Klein, Keelin Murphy, Marius Staring, and Josien~PW Pluim.
\newblock A survey of medical image registration--under review, 2016.

\bibitem[Wang et~al.(2023)Wang, Evan, Dalca, and Sabuncu]{wang2023robust}
Alan~Q Wang, M~Yu Evan, Adrian~V Dalca, and Mert~R Sabuncu.
\newblock A robust and interpretable deep learning framework for multi-modal registration via keypoints.
\newblock \emph{Medical Image Analysis}, 90:\penalty0 102962, 2023.

\bibitem[Xiao et~al.(2024)Xiao, Liu, Tang, Liu, Wu, and Zhao]{xiao2024reduction}
Chun-Lin Xiao, Lu-Lin Liu, Wen Tang, Wu-Yang Liu, Long-Yan Wu, and Kai Zhao.
\newblock Reduction of the trans-cortical vessel was associated with bone loss, another underlying mechanism of osteoporosis.
\newblock \emph{Microvascular Research}, 152:\penalty0 104650, 2024.

\bibitem[Xiao et~al.(2022)Xiao, Cui, Hsu, Peng, Jiang, Xu, Ma, Liu, and Lu]{xiao2022global}
P-L Xiao, A-Y Cui, C-J Hsu, R~Peng, N~Jiang, X-H Xu, Y-G Ma, D~Liu, and H-D Lu.
\newblock Global, regional prevalence, and risk factors of osteoporosis according to the world health organization diagnostic criteria: a systematic review and meta-analysis.
\newblock \emph{Osteoporosis International}, 33\penalty0 (10):\penalty0 2137--2153, 2022.

\bibitem[Xiong et~al.(2021)Xiong, Yi, Lin, Qiu, Shu, and Zhang]{xiong2021evaluation}
Zhencheng Xiong, Ping Yi, Jialiang Lin, Shengfeng Qiu, Li~Shu, and Chi Zhang.
\newblock Evaluation of the efficacy of stem cell therapy in ovariectomized osteoporotic rats based on micro-ct and dual-energy x-ray absorptiometry: A systematic review and meta-analysis.
\newblock \emph{Stem Cells International}, 2021\penalty0 (1):\penalty0 1439563, 2021.

\bibitem[Yang et~al.(2015)Yang, Li, Campbell, and Jia]{yang2015go}
Jiaolong Yang, Hongdong Li, Dylan Campbell, and Yunde Jia.
\newblock Go-icp: A globally optimal solution to 3d icp point-set registration.
\newblock \emph{IEEE transactions on pattern analysis and machine intelligence}, 38\penalty0 (11):\penalty0 2241--2254, 2015.

\bibitem[Yousefzadeh et~al.(2020)Yousefzadeh, Kashfi, Jeddi, and Ghasemi]{yousefzadeh2020ovariectomized}
Nasibeh Yousefzadeh, Khosrow Kashfi, Sajad Jeddi, and Asghar Ghasemi.
\newblock Ovariectomized rat model of osteoporosis: a practical guide.
\newblock \emph{EXCLI journal}, 19:\penalty0 89, 2020.

\bibitem[Zhou et~al.(2018)Zhou, Park, and Koltun]{open3d}
Qian-Yi Zhou, Jaesik Park, and Vladlen Koltun.
\newblock Open3d: A modern library for 3d data processing.
\newblock \emph{arXiv preprint arXiv:1801.09847}, 2018.

\end{thebibliography}





\end{document}